\pdfoutput=1
\documentclass[11pt]{article}

\usepackage[preprint]{acl}

\usepackage{times}
\usepackage{latexsym}
\usepackage{amsmath}
\usepackage[T1]{fontenc}

\usepackage[utf8]{inputenc}

\usepackage{microtype}

\usepackage{inconsolata}

\usepackage{graphicx}
\usepackage{xcolor}

\usepackage{booktabs}
\usepackage{tabularx}
\usepackage{array}
\usepackage{multirow}
\usepackage{multicol}
\usepackage{pifont}

\newcommand{\cmark}{\ding{51}}
\newcommand{\xmark}{\ding{55}}

\title{Multi-SFT: Permissively-Licensed Multilingual Instruction-Following}

\author{
  \textbf{Daniela Baiamonte\textsuperscript{1}},
  \textbf{Kamyar Zeinalipour\textsuperscript{2,3}},
  \textbf{Elena Fano\textsuperscript{1}},
  \textbf{Stefano Simonazzi\textsuperscript{1}},
  \textbf{Andrea Zugarini\textsuperscript{1}}
\\
  \textsuperscript{1}villanova.ai,\quad \textsuperscript{2}Yukai!,\quad \textsuperscript{3}University of Siena
\\
  \small{
    \textbf{Correspondence:} \href{mailto:kamyar.zeinalipour2@unisi.it}{\texttt{kamyar.zeinalipour2@unisi.it}}
  }
}

\begin{document}
\maketitle
\begin{abstract}
Supervised fine-tuning datasets remain overwhelmingly English-centric and commercially restricted, as their content is frequently generated by proprietary models whose terms of service forbid training competing systems. We introduce \textit{Multi-SFT}, a commercially permissive collection of 502,818 instruction-response pairs across five European languages, built via \textit{translation} and \textit{regeneration} pipelines for tasks where translation is insufficient. Human evaluation yields inter-annotator agreement of at least 0.87 across all datasets and languages. Fine-tuning nine models from three different families (EuroLLM, OLMo 2, Qwen3; 1B--32B) and evaluating them on 13 benchmarks shows that \textit{Multi-SFT} alone is competitive with larger augmented mixes and official instruction-tuned releases, with decisive advantages on safety tasks.

\end{abstract}

\section{Introduction}
Supervised fine-tuning (SFT) is central to aligning pretrained LLMs with human instructions \citep{wei2021finetuned, ouyang2022training, taori2023alpaca}, with data quality, diversity, and scale being critical for downstream behavior \citep{wang2023self, zhou2023lima}. Yet, most existing SFT datasets remain English-centric and are affected by a less visible but equally serious problem: licensing. Many corpora carry permissive labels (MIT, Apache 2.0, Creative Commons), while their content was generated by proprietary models such as GPT-4 or Gemini, whose terms of service explicitly prohibit use for training competing models.
\begin{table}[t]
    \centering
    \small
    \begin{tabular}{cccc}
    \toprule
         \textbf{Project} &  \textbf{Multilingual} &  \textbf{\shortstack{Dataset\\License}} & \textbf{\shortstack{Data\\Provenance}}\\
         \midrule
         Apertus & \cmark & \cmark & \xmark \\
         EuroLLM & \cmark & \textasciitilde & \xmark \\
         Nemotron & \xmark & \cmark & \cmark \\
         OLMo & \xmark & \cmark & \xmark \\
         Salamandra & \cmark & \xmark & \xmark \\
         Multi-SFT & \cmark & \cmark & \cmark \\
         \bottomrule
    \end{tabular}
    \caption{\textbf{Comparison of SFT mixtures used in recent open LLM projects.}
\cmark~denotes full compliance, \xmark~denotes non-compliance, and \textasciitilde~denotes partial compliance. \textit{Dataset License} indicates whether all training data is released under permissive licenses. \textit{Data Provenance} indicates whether synthetic data is generated exclusively with permissive or open-weight models, excluding proprietary sources such as GPT-4 or Gemini.}
    \label{tab:sftcomparison}
\end{table}

We introduce \textit{Multi-SFT}, a collection of 502,818 instruction-response pairs spanning five European languages (English, Italian, French, Spanish, and German), aggregating 13 heterogeneous datasets built exclusively from permissively licensed data and models (Table~\ref{tab:sftcomparison}). The collection is constructed via two pipelines: a \textit{translation} pipeline that adapts existing English-centric datasets into the four target languages through an expert-driven, quality-assured process, and a \textit{regeneration} pipeline that synthesizes multilingual content from scratch for tasks where translation is insufficient, such as conversational interaction, constrained instruction following, and safety. Both pipelines undergo systematic human evaluation by native or near-native annotators, yielding inter-annotator agreement of at least 0.87 across all datasets and languages.

We validate \textit{Multi-SFT} by fine-tuning nine models from three families (EuroLLM, OLMo 2, Qwen3; 1B--32B parameters) on 13 benchmarks.
Our experiments show that the multilingual, permissively-licensed collection achieves stronger performance on non-English languages than the original English sources, which contain non-permissive data from closed LLMs, while surprisingly maintaining performance on English benchmarks as well. Furthermore, in multilingual settings, Multi-SFT outperforms official instruct models in most cases and stays competitive on English.
Multi-SFT is publicly available.\footnote{\url{https://huggingface.co/collections/VillanovaAI/multi-sft-6ad0e9ea53f7d008611b8204}}

\begin{table*}
\resizebox{\textwidth}{!}{
\begin{tabular}{l l c l r r r r r r}
\toprule
Dataset & Domain & Strategy & License & EN & DE & ES & FR & IT & Total \\
\midrule
\textit{Multi-AdvBench}            & Safety                & Regeneration   & MIT          & 104 & 104 & 104 & 104 & 104 & 520 \\
\textit{Multi-AyaRedteaming}       & Safety                & Regeneration   & Apache 2.0   & 812 & 835 & 330 & 754 & 913 & 3,644 \\
\textit{multi-dialogues}           & Conversation          & Regeneration   & ODC-BY-1.0   & - & 12,001 & 11,999 & 12,000 & 12,001 & 48,001 \\
\textit{multi-dolly-15k}           & General QA            & Translation  & CC BY-SA 3.0 & 6,004 & 2,250 & 2,251 & 2,252 & 2,222 & 14,979 \\
\textit{Multi-FLAN}                & Reasoning             & Translation  & Apache 2.0   & 36,069 & 13,495 & 13,441 & 13,492 & 13,485 & 89,982 \\
\textit{multi-mentorES}            & General QA            & Translation  & CC BY 4.0     & - & 2,541 & 2,552 & 2,534 & 2,548 & 10,175 \\
\textit{Multi-PersonaIF}           & Instruction Following & Regeneration   & ODC-BY-1.0   & 15,000 & 15,000 & 15,000 & 15,000 & 15,000 & 75,000 \\
\textit{Multi-RAG-Multilingual}    & General QA            & Translation  & CC-BY-SA-4.0 & 4,999 & 4,999 & 5,631 & 4,998 & 5,632 & 26,259 \\
\textit{multi-safety}              & Safety                & Regeneration   & ODC-BY-1.0   & 250 & 250 & 250 & 250 & 250 & 1,250 \\
\textit{Multi-SciRIFF}             & Scientific Literature & Translation  & ODC-BY       & 28,560 & 10,521 & 10,508 & 10,540 & 10,392 & 70,521 \\
\textit{multi-smol\_rewrite}       & Rephrasing            & Translation  & Apache 2.0   & 21,305 & 7,990 & 7,989 & 7,988 & 7,990 & 53,262 \\
\textit{multi-smol\_summarize}     & Summarization         & Translation + Regeneration & Apache 2.0    & 38,424 & 14,409 & 14,402 & 14,410 & 14,409 & 96,054 \\
\textit{Multi-TableGPT}            & Table Understanding   & Translation  & MIT          & 5,279 & 1,964 & 1,969 & 1,980 & 1,979 & 13,171 \\
\midrule
\textbf{Total} & & & & & & & & & \textbf{502,818} \\

\bottomrule
\end{tabular}
}
\caption{Overview of the \textit{multi-} dataset collection. Source datasets and pipelines: AdvBench~\citep{zou2023universal}, Aya Red-teaming~\citep{ahmadian2024multilingual}, SODA~\citep{kim2023soda}, Dolly~\citep{conover2023free}, FLAN~\citep{longpre2023flan}, Persona IF~\citep{lambert2024tulu}, RAG Multilingual~\citep{projecteaina2026rag}, SciRIFF~\citep{wadden2025sciriff}, SmolTalk~\citep{Allal2025SmolLM2WS}, and Table-GPT~\citep{li2024table}.}
\label{tab:datasets}
\end{table*}

\section{Related Work}
Supervised fine-tuning (SFT) adapts pre-trained language models to follow instructions via curated prompt-response pairs~\cite{ouyang2022training}. English SFT resources span a continuum from fully human-written data---such as FLAN~\cite{longpre2023flan}, Dolly~\cite{conover2023free}, and Open Assistant~\cite{kopf2023openassistant}---to hybrid pipelines combining human annotation with synthetic generation, as in Tülu 3~\cite{lambert2024tulu}. Recent large-scale synthetic efforts such as Nemotron~\cite{blakeman2025nvidia} and Smoltalk~\cite{Allal2025SmolLM2WS} demonstrate the scalability of automatically constructed instruction datasets while explicitly addressing licensing and provenance through permissively licensed sources and models.
High-quality multilingual SFT resources for European languages remain comparatively limited, and existing datasets often face restricted licensing or provenance concerns due to reliance on proprietary generative models. The Aya initiative~\cite{singh2024aya} provides the broadest multilingual effort to date, though its wide coverage results in limited per-language depth and variable quality in automatically generated portions. EuroLLM~\cite{martins2024eurollm}, Tülu 3~\cite{lambert2024tulu}, Salamandra~\cite{gonzalez2025salamandra}, and Apertus~\cite{apertus2025apertus} each advance multilingual or open post-training in different respects, yet all include components generated or processed with proprietary models, limiting their suitability for provenance-constrained training.
The Smoltalk multilingual extension represents one of the few systematic attempts to construct permissively licensed multilingual SFT data through translation and open-model generation. Our work follows a similar philosophy while expanding task coverage, introducing expert-driven quality validation, and explicitly addressing provenance constraints through selective regeneration of non-compliant examples.

\section{Multi-SFT}
Multi-SFT is a supervised fine-tuning collection of 13 datasets covering five European languages: English, Italian, French, Spanish, and German. Overall, it contains 502,818 diverse examples (e.g., QA, instruction following). Per-dataset details are outlined in Table~\ref{tab:datasets}.

Many existing SFT datasets, despite declaring permissive licenses, are built on human-authored prompts paired with responses generated by proprietary models, which restricts their use in commercial settings. Therefore, we erased such content and regenerated it using open-weight, permissively licensed LLMs. Depending on the source dataset, we employ a \textit{translation} pipeline, where existing (English-centric) datasets are converted into multilingual ones, or a \textit{regeneration} process, where synthetic content is re-generated from scratch with license-compliant language models.

\subsection{Dataset selection criteria}
Rather than aggregating large volumes of generic instruction-response pairs, we focus on diversity and high quality. Thus, source datasets were selected to cover a vast and heterogeneous mixture of competences and tasks (see Table~\ref{tab:datasets}).

\paragraph{Licensing and Compliance.}
\label{sec:compliance}
A prerequisite for inclusion in the collection was the availability of a permissive license compatible with commercial use.
Moreover, we also take into account the provenance of synthetic content present in the datasets. This aspect is often overlooked, but proprietary language models such as GPT, Claude or Gemini restrict or prohibit their use for the development of commercial models. Hence, we excluded all datasets where there is evidence that entries were generated using models with such restrictive licensing, regardless of the license applied to the dataset itself.
We consider as "evidence" of non-permissive model use any dataset mentioning in its card, git repository or paper the use of commercial models.
Whenever a dataset is released under a permissive license but a portion of its entries were generated using a restricted model, we opted for selective regeneration: the affected entries were regenerated using permissively licensed open-weight models, specifically: Mistral-small\footnote{\texttt{mistralai/Mistral-Small-3.1-24B-Instruct-2503}}, Qwen3-VL-235B-A22B\footnote{\texttt{Qwen/Qwen3-VL-235B-A22B-Instruct}}, and Qwen3.5-122B-A10B\footnote{\texttt{Qwen/Qwen3.5-122B-A10B}}. These models were selected based on their empirical performance across translation and generation tasks, and their licensing compatibility with downstream training (see Appendix~\ref{appx:model_selection} for details).
Out of 59 candidate datasets initially considered (excluding math and coding topics), 45 were excluded due to licensing or provenance violations (see Appendix~\ref{appx:dataset_selection}). A sample of regenerated examples is provided in Appendix~\ref{appx:provenance_audit}.

\subsection{Translation Pipeline}
\label{sec:translation}
Translated datasets were produced through an iterative, expert-driven pipeline. For each target language, a native or near-native linguist independently developed a translation prompt, selecting the generation model and temperature in collaboration with the team. An initial batch of samples was generated and evaluated blind by a separate QA reviewer with equivalent language expertise. Based on the reviewer's feedback, the prompt, model, or parameters were revised and the process repeated until the output quality was deemed satisfactory.

A special case within the translation pipeline is \textit{multi-smol\_summarize}, which combines translation and generation. In the original Smoltalk summarization dataset, source documents are paired with summaries generated from a very small set of instruction variants. In our version, the source documents are first translated into the target languages; the summaries are then generated from scratch using the models listed in Section \ref{sec:compliance}. Crucially, we expanded the original instruction set by introducing additional variants expressing formal output requirements, such as producing the summary as a bulleted list or as a Markdown table, thereby increasing both the linguistic and structural diversity of the dataset.

\subsection{Synthetic Data Generation}
Five datasets in the collection were constructed through synthetic generation. For each, the same expert validation process described in Section \ref{sec:translation} was applied: prompt design, blind evaluation, and iterative refinement by language-expert reviewers.
\paragraph{Multi-AdvBench.} The original multilingual AdvBench dataset consists of harmful prompt--unsafe response pairs. We retained the prompts and regenerated all responses using \texttt{Qwen3-VL-235B-A22B-Instruct}, with the explicit objective of producing safe, refusal-oriented completions. To avoid any overlap across languages, the original 520 prompts were randomly partitioned into five disjoint blocks of 104, one per language, so that each prompt appears in exactly one language. All generated pairs were manually reviewed.
\paragraph{Multi-AyaRedTeaming.} Starting from the English, French, and Spanish subsets of the Aya Red-teaming dataset, we repurposed this adversarial resource for SFT by transforming harmful prompts into instruction-following examples with safe target responses. We generated the expected safe responses using \texttt{Qwen3-VL-235B-A22B-Instruct}. For Italian and German, where no source data was available, we used the English prompts as seeds: the same model was prompted to produce culturally appropriate questions in the target language and then to generate the corresponding safe responses.
All examples were manually reviewed and corrected.
\paragraph{Multi-Dialogues.} It consists of synthetically generated multi-turn everyday conversations, produced by adapting the SODAverse pipeline from \citet{kim2023soda}. The original SODA dataset was built from social common-sense triples extracted from the \textsc{Atomic}$^{10\mathrm{x}}$ knowledge graph \cite{west2022symbolic}: each triple is first expanded into a short narrative, which is then used to prompt a model to generate a dialogue. We replicated the pipeline for French, German, Italian, and Spanish using \texttt{Mistral-Small-3.1-24B-Instruct-2503}, generating 12,000 dialogues per language. Then, \texttt{Qwen3-VL-235B-A22B-Instruct} was used for cleaning, formatting, and profanity filtering; American characters were replaced with culturally-appropriate names for each target language.
\paragraph{Multi-PersonaIF.} This dataset targets instruction following under explicit constraints, following the methodology of AllenAI's Persona IF \cite{lambert2024tulu}. Instructions are generated by randomly sampling 1--3 constraints from a predefined list derived from IFEval \cite{zhou2023ifeval}, using persona descriptions from a multilingual extension of PersonaHub \cite{ge2406scaling} to diversify instruction styles. Responses are generated by the same model immediately after instruction generation. We replicated the full pipeline using \texttt{Qwen3.5-122B-A10B} and translated and adapted all prompts, seed instructions, and constraint definitions for the five target languages. One methodological modification was introduced: the original constraint sampling procedure can yield logically incompatible combinations (e.g., write in lowercase and include a minimum number of capitalized words). We revised the sampling logic to prevent such conflicts.
\paragraph{Multi-Safety.} A small dataset of harmful request--safe response pairs was constructed internally: harmful prompts were authored by team members, and safe responses were generated using \texttt{Qwen3-VL-235B-A22B-Instruct}. All pairs were manually reviewed before inclusion.

\subsection{Annotation Process}
To assess the quality of translated and generated datasets, we conducted a systematic human evaluation covering all datasets produced through the translation and generation pipelines.
Inspired by \citet{wang2023self}, instruction-response pairs were annotated on an A-to-D grading scale (Table~\ref{tab:annotation_schema}).
Full details are reported in Appendix~\ref{appx:annotation_design}.
\begin{table}[t]
\centering
\footnotesize

\begin{tabularx}{\columnwidth}{
>{\raggedright\arraybackslash}X
ccccc
}

\toprule
\textbf{Dataset} & \textbf{DE} & \textbf{EN} & \textbf{ES} & \textbf{FR} & \textbf{IT} \\
\midrule
         \textit{multi-dialogues} & 98 & -- & 99 & 100 & 95 \\
         \textit{multi-dolly-15k} & 100 & -- & 93 & 100 & 94 \\
         \textit{Multi-FLAN} & 90 & -- & 87 & 84 & 88 \\
         \textit{multi-mentorES} & 95 & -- & -- & 96 & 93 \\
         \textit{Multi-PersonaIF} & 87 & 89 & 83 & 80 & 72 \\
         \textit{Multi-RAG-Multilingual} & 100 & -- & -- & 96 & 90 \\
         \textit{Multi-SciRIFF} & 97 & -- & 97 & 92 & 87 \\
         \textit{multi-smol\_rewrite} & 100 & -- & 97 & 99 & 98 \\
         \textit{multi-smol\_summarize} & 99 & 94 & 94 & 96 & 95 \\
         \textit{Multi-TableGPT} & 98 & -- & 76 & 97 & 98 \\
    \bottomrule
\end{tabularx}

\caption{Percentage of positively-rated entries by dataset and language. Values represent the average of the two independent annotators.}
\label{tab:annotations_dist}
\end{table}

For each dataset and each language, a sample of 50 entries was selected for annotation, with one exception: for Multi-FLAN, which aggregates four sub-datasets (CoT, P3, FLAN2021, NIv2), 60 entries were annotated per language, 15 per sub-dataset, to ensure proportional coverage. Each sample was annotated independently by two annotators, both with university-level expertise in the target language (advanced students or graduates in linguistics or related fields).
English annotation was performed only for datasets where English content was generated from scratch, i.e., Multi-PersonaIF and multi-smol\_summarize.

\subsection{Quality Assessment}
\paragraph{Metrics.} Inter-annotator agreement was computed using Gwet's AC2 \cite{Gwet2008} with quadratic weights on the collapsed three-way label set (A/B vs.\ C vs.\ D). AC2 was preferred over Cohen’s $\kappa$ \cite{Cohen1960} because the annotation distributions are strongly skewed toward the positive category (A/B), a condition under which $\kappa$ is known to produce paradoxically low values even when observed agreement is high. The quadratic weighting scheme is appropriate for ordinal scales, where the distance between non-adjacent categories (e.g., A/B and D) should be penalised more heavily than disagreements between adjacent ones.

\paragraph{Results.} Agreement results are reported in Table~\ref{tab:agreement}. The distribution of raw annotation labels across datasets and languages is provided in Table~\ref{tab:annotations_dist}. Agreement is consistently high across datasets and languages, with values exceeding 0.90 in the majority of cases. This indicates reliable interpretation of the annotation schema and stable quality signals. Three datasets show lower but still substantial agreement: Multi-FLAN (FR: 0.93, DE: 0.90), Multi-PersonaIF (FR: 0.88), and Multi-TableGPT (ES: 0.87), reflecting known task-inherent challenges discussed in Appendix~\ref{app:agreement_analysis}.
The label distribution (Table~\ref{tab:annotations_dist}) reveals that the vast majority of entries fall in the positive category, confirming overall pipeline quality. Multi-PersonaIF exhibits the highest proportion of badly-rated entries (up to 28\% in Italian), stemming from two sources: constraint violations and incomplete instructions lacking necessary context. While constraint violations reflect inherent task difficulty, incomplete instructions represent actionable quality issues addressable through improved prompt engineering and post-hoc filtering.

\begin{table}[t]
\centering
\footnotesize
\setlength{\tabcolsep}{3pt}

\begin{tabularx}{\columnwidth}{
>{\raggedright\arraybackslash}X
ccccc
}

\toprule
\textbf{Dataset} & \textbf{DE} & \textbf{EN} & \textbf{ES} & \textbf{FR} & \textbf{IT} \\
\midrule
         \textit{multi-dialogues} & 0.99 & -- & 0.99 & 1 & 0.97 \\
         \textit{multi-dolly-15k} & 1 & -- & 0.95 & 1 & 0.98 \\
         \textit{Multi-FLAN} & 0.90 & -- & 0.97 & 0.93 & 0.93 \\
         \textit{multi-mentorES} & 0.99 & -- & -- & 0.96 & 0.98 \\
         \textit{Multi-PersonaIF} & 0.98 & 0.94 & 0.98 & 0.88 & 0.94 \\
         \textit{Multi-RAG-Multilingual} & 1 & -- & -- & 0.98 & 0.97 \\
         \textit{Multi-SciRIFF} & 0.98 & -- & 0.97 & 0.96 & 0.94 \\
         \textit{multi-smol\_rewrite} & 1 & -- & 0.97 & 0.99 & 0.99 \\
         \textit{multi-smol\_summarize} & 0.99 & 0.97 & 0.97 & 0.98 & 0.97 \\
         \textit{Multi-TableGPT} & 0.99 & -- & 0.87 & 0.98 & 0.99 \\
    \bottomrule
\end{tabularx}

\caption{Inter-annotator agreement (Gwet's AC2) by dataset and language.}
\label{tab:agreement}
\end{table}

\section{Experiments}
\subsection{Experimental Setup}
We pick nine \textbf{base} models from three families: EuroLLM~\cite{martins2024eurollm} (1.7B, 9B, 22B), OLMo~2~\cite{OLMo20242O2} (1B, 7B, 32B), and Qwen3~\cite{Yang2025Qwen3TR} (1.7B, 8B, 14B), overall spanning from 1B to 32B parameters.
These three families differ significantly in their nature. Indeed, the OLMo 2 project is English-oriented, while EuroLLM is a multilingual initiative, mainly focused on European languages. Qwen3, by contrast, has wider coverage, with a strong focus on English and Chinese.

To prove the effectiveness of our multilingual permissive collection, we train each base model on three different corpora. First, we consider ours, Multi-SFT (502,818 examples), denoted as \textsc{ml}. In \textsc{full}, we expand the training set of \textsc{ml} to 1.6M examples, by adding existing open-source datasets: the Aya Collection~\cite{singh2024aya} (928K), multilingual SmolTalk~\cite{Allal2025SmolLM2WS} (127K), GSM8K~\cite{Cobbe2021TrainingVT} (37K), and RAG Multilingual~\cite{projecteaina2026rag} (10K).
Finally, we include a baseline (\textsc{bl}) trained on the same source datasets as \textsc{ml} but kept in their original English form, to isolate the effect of multilingual content.
As a reference of comparison we also evaluated the officially released instruct models, here denoted as \textsc{sft$^\dagger$}.
Training details are reported in Appendix~\ref{appx:training_details}.

\begin{table*}[t]
\centering
\setlength{\tabcolsep}{3.5pt}
\tiny
\renewcommand{\arraystretch}{1.05}
\begin{tabular}{cl cccccccc c ccccc c}
\toprule
& & \multicolumn{9}{c}{\textbf{Multilingual}} & \multicolumn{6}{c}{\textbf{English}} \\
\cmidrule(lr){3-11} \cmidrule(lr){12-17}
\textbf{Family} & \textbf{Model} & \textbf{MARCO} & \textbf{FLORES} & \textbf{Belebele} & \textbf{XQuAD} & \textbf{M-ALERT} & \textbf{MMLU} & \textbf{ARC} & \textbf{XStory} & \textbf{Avg} & \textbf{XQuAD} & \textbf{DROP} & \textbf{IFE} & \textbf{MMLU} & \textbf{ARC} & \textbf{Avg} \\
\midrule
\multirow{12}{*}{\rotatebox{90}{EuroLLM}}
 & 1.7B \textsc{ml} & \textbf{31.7} & 49.0 & 23.4 & 33.4 & 43.6 & 24.5 & 32.0 & 58.4 & \textbf{37.0} & 32.8 & 23.0 & 32.3 & 23.5 & 34.0 & 29.1 \\
 & 1.7B \textsc{full} & 21.1 & 45.3 & 23.0 & 25.7 & 21.6 & 24.4 & 32.2 & \textbf{61.2} & 31.8 & 24.6 & 21.1 & 27.9 & 23.4 & 34.1 & 26.2 \\
  & 1.7B \textsc{bl} & 27.1 & 52.2 & \textbf{25.9} & \textbf{42.6} & 29.0 & 24.5 & 32.1 & 58.6 & 36.5 & \textbf{42.5} & \textbf{25.2} & \textbf{39.9} & 24.3 & \textbf{37.5} & \textbf{33.9} \\
 & 1.7B \textsc{sft$^\dagger$} & 16.5 & \textbf{59.1 }& 25.4 & 25.0 & \textbf{44.4} & \textbf{26.3}& \textbf{33.3} & 60.8 & 36.4 & 37.1 & 25.1 & 20.3 & \textbf{26.7} & 37.4 & 29.3 \\
\cmidrule(l){2-17}
 & 9B \textsc{ml} & \textbf{50.2} & 62.0 & 64.6 & 75.8 & \textbf{56.3} & \textbf{50.7} & \textbf{48.9} & 64.5 & \textbf{59.1} & 73.9 & 46.6 & \textbf{59.5} & 53.0 & \textbf{52.9} & \textbf{57.2} \\
 & 9B \textsc{full} & 25.1 & 57.8 & 62.3 & 76.2 & 33.9 & 44.1 & 43.3 & 57.4 & 50.0 & \textbf{75.6} & \textbf{48.1} & 26.4 & 46.8 & 44.5 & 48.3 \\
  & 9B \textsc{bl} & 41.7 & 60.5 & 64.8 & \textbf{76.6} & 36.3 & 46.5 & 43.0 & 64.5 & 54.2 & 75.4 & 47.4 & 52.9 & 50.2 & 46.7 & 54.5 \\
 & 9B \textsc{sft$^\dagger$} & 42.3 & \textbf{62.8} & \textbf{70.6} & 68.5 & 55.7 & 45.4 & 44.9 & \textbf{67.2} & 57.2 & 64.5 & 40.5 & 57.5 & \textbf{54.5} & 50.3 & 53.5 \\
\cmidrule(l){2-17}
 & 22B \textsc{ml} & 49.4 & 62.5 & 77.9 & \textbf{79.4} & 53.7 & 57.5 & \textbf{51.7} & 66.7 & \textbf{62.4} & \textbf{77.3} & 52.0 & 59.3 & 60.4 & 55.3 & 60.9 \\
 & 22B \textsc{full} & 47.3 & 62.2 & 75.0 & 76.6 & \textbf{56.0} & 56.0 & 48.9 & 68.2 & 61.3 & 76.8 & 47.0 & 58.0 & 59.2 & 53.6 & 58.9 \\
  & 22B \textsc{bl} & 46.4 & 62.3 & 77.7 & 76.3 & 37.9 & 56.3 & 50.8 & 67.2 & 59.4 & 76.2 & \textbf{55.2} & 63.0 & 60.7 & \textbf{57.0} & \textbf{62.4} \\
 & 22B \textsc{sft$^\dagger$} & \textbf{50.2} & \textbf{63.5} & \textbf{79.2} & 70.5 & 40.8 & \textbf{60.9} & 47.3 & \textbf{69.1} & 60.2 & 67.6 & 49.6 & \textbf{70.1} & \textbf{63.7} & 52.9 & 60.8 \\
\midrule
\multirow{12}{*}{\rotatebox{90}{OLMo~2}}

  & 1B \textsc{ml} & \textbf{32.9} & 44.6 & \textbf{33.3} & 26.4 & \textbf{49.5} & \textbf{31.0} & 26.9 & 55.3 & 37.5 & 24.8 & 33.6 & 41.2 & 39.2 & 38.1 & 35.4 \\

 & 1B \textsc{full} & 18.5 & 40.8 & 28.7 & 47.0 & 30.8 & 28.1 & 26.6 & 54.9 & 34.4 & 41.3 & \textbf{34.3} & 22.6 & 34.3 & 38.5 & 34.2 \\

    & 1B \textsc{bl} & 21.1 & 42.6 & 32.0 & 47.0 & 21.7 & 30.1 & 26.2 & 53.6 & 34.3 & \textbf{51.2} & 33.4 & 46.2 & \textbf{40.4} & \textbf{38.7} & 42.0 \\

  & 1B \textsc{sft$^\dagger$} & 30.6 & \textbf{51.2} & 31.1 & \textbf{48.6} & 19.1 & 30.3 & \textbf{30.6} & \textbf{59.7} & \textbf{37.6} & 50.9 & 31.7 & \textbf{66.0} & 34.0 & 38.2 & \textbf{44.2} \\

\cmidrule(l){2-17}

  & 7B \textsc{ml} & \textbf{47.8} & 46.7 & \textbf{69.8} & 50.8 & 51.8 & 44.6 & 36.3 & 60.4 & 51.0 & 56.6 & 58.4 & 56.0 & 57.5 & 50.8 & 55.9 \\

  & 7B \textsc{full} & 45.0 & 51.9 & 67.8 & \textbf{65.2} & \textbf{56.9} & \textbf{46.2} & \textbf{38.2} & 63.3 & \textbf{54.3} & \textbf{67.1} & 55.0 & 55.3 & \textbf{58.3} & \textbf{51.8} & 57.5 \\

    & 7B \textsc{bl} & 34.0 & 44.6 & 64.4 & 61.2 & 32.5 & 42.2 & 34.3 & 59.4 & 46.6 & 62.3 & 56.6 & 61.6 & 58.0 & 50.9 & 57.9 \\

     & 7B \textsc{sft$^\dagger$} & 43.4 & \textbf{57.5} & 66.6 & 63.7 & 41.0 & 44.7 & 37.9 & \textbf{64.6} & 52.4 & 67.0 & \textbf{58.9} & \textbf{71.9} & 57.3 & 47.4 & \textbf{60.5} \\

\cmidrule(l){2-17}

  & 32B \textsc{ml} & 52.1 & 29.8 & 82.2 & 45.3 & \textbf{55.4} & 61.2 & 50.6 & 66.6 & 55.4 & 51.6 & 54.3 & 67.7 & 69.2 & 59.2 & 60.4 \\

 & 32B \textsc{full} & 49.2 & 46.9 & 85.8 & 71.5 & 53.5 & 62.7 & 50.2 & 64.5 & 60.5 & \textbf{72.5} & \textbf{72.7} & 59.7 & 69.3 & 56.4 & 66.1 \\

  & 32B \textsc{bl} & 42.6 & 32.7 & 77.7 & \textbf{73.1} & 41.1 & 58.2 & 44.5 & 66.4 & 54.5 & 72.4 & 62.4 & 64.7 & 67.4 & 59.3 & 65.2 \\

 & 32B \textsc{sft$^\dagger$} & \textbf{58.6} & \textbf{61.1} & \textbf{87.2} & 69.9 & 49.7 & \textbf{66.1} & \textbf{55.8} & \textbf{76.4} & \textbf{65.6} & 67.0 & 71.6 & \textbf{82.3} & \textbf{72.9} & \textbf{66.0} & \textbf{72.0} \\

\midrule
\multirow{12}{*}{\rotatebox{90}{Qwen3}}
 & 1.7B \textsc{ml} & 43.4 & 51.6 & 74.1 & 45.4 & 55.4 & 51.0 & 35.7 & 59.9 & 52.1 & 50.0 & 46.6 & 53.8 & 59.0 & 41.8 & 50.2 \\

 & 1.7B \textsc{full} & 44.7 & 52.5 & 72.2 & 53.6 & \textbf{57.6} & \textbf{52.2} & \textbf{41.6} & \textbf{60.0} & \textbf{54.3} & 60.0 & \textbf{47.2} & 52.1 & 58.3 & \textbf{52.8} & 54.1 \\

  & 1.7B \textsc{bl} & 40.3 & 50.0 & \textbf{74.5} & \textbf{65.9} & 42.4 & 51.6 & 37.1 & 59.9 & 52.7 & \textbf{65.8} & 46.5 & 58.2 & \textbf{59.2} & 44.3 & \textbf{54.8} \\

   & 1.7B \textsc{sft$^\dagger$} & \textbf{46.0} & \textbf{54.7} & 36.2 & 61.7 & 30.7 & 31.8 & 32.1 & 55.1 & 43.5 & 61.4 & 37.7 & \textbf{67.8} & 50.9 & 33.4 & 50.2 \\

\cmidrule(l){2-17}
 & 8B \textsc{ml} & 50.9 & 50.8 & \textbf{89.2} & 62.1 & 53.8 & \textbf{67.9} & 51.0 & 64.0 & 61.2 & 66.2 & \textbf{64.0} & 69.5 & 72.8 & 57.8 & \textbf{66.1} \\

 & 8B \textsc{full} & 50.6 & 55.3 & 88.5 & 65.1 & \textbf{55.6} & 67.2 & \textbf{51.3} & 64.1 & \textbf{62.2} & \textbf{70.9} & 57.4 & 57.7 & 71.6 & \textbf{59.6} & 63.4 \\

 & 8B \textsc{bl} & 49.5 & 50.4 & 86.4 & \textbf{72.0} & 42.7 & 67.4 & 50.8 & \textbf{65.1} & 60.5 & \textbf{70.9} & 59.1 & 67.5 & \textbf{73.2} & 57.0 & 65.5 \\

 & 8B \textsc{sft$^\dagger$} & \textbf{57.7} & \textbf{59.7} & 28.4 & 71.3 & 41.3 & 33.0 & 37.8 & 61.7 & 48.9 & 70.3 & 51.4 & \textbf{81.3} & 32.1 & 41.9 & 55.4 \\

\cmidrule(l){2-17}
 & 14B \textsc{ml} & 52.9 & 51.8 & \textbf{91.7} & 61.8 & \textbf{55.2} & \textbf{72.6} & \textbf{54.5} & 65.3 & 63.2 & 59.8 & \textbf{71.7} & 69.5 & \textbf{77.4} & 59.9 & 67.7 \\

 & 14B \textsc{full} & 52.9 & 51.5 & 91.4 & 63.4 & 54.6 & 71.0 & 54.1 & \textbf{67.2} & 63.3 & 65.4 & 62.5 & 63.2 & 75.2 & \textbf{61.7} & 65.6 \\

  & 14B \textsc{bl} & 51.1 & 50.1 & 91.3 & \textbf{73.3} & 51.3 & 71.9 & 52.3 & 67.0 & \textbf{63.5} & \textbf{72.3} & 66.1 & 70.8 & 77.2 & 60.8 & \textbf{69.4} \\

 & 14B \textsc{sft$^\dagger$} & \textbf{60.0} & \textbf{60.8} & 23.7 & 71.1 & 36.1 & 25.8 & 38.7 & 62.3 & 47.3 & 67.0 & 52.5 & \textbf{85.0} & 23.4 & 42.7 & 54.1 \\

\bottomrule
\end{tabular}
\caption{Benchmark results divided between \textbf{Multilingual} (left) and \textbf{English} (right). \textsc{ml}: Multi-SFT ; \textsc{full}: extended mix; \textsc{bl}: baseline SFT; \textsc{sft$^\dagger$}: official provider model. Average performance is given per group (see \textbf{Avg} columns).}
\label{tab:all_results}
\end{table*}

\paragraph{Evaluation benchmarks.}
We test all models on 10 distinct benchmarks, yielding 13 benchmarks in total since three benchmarks (XQuAD, M-MMLU, ARC) are evaluated in both multilingual and English settings. The multilingual ones (Table~\ref{tab:all_results}) are: MARCO~\cite{zeng2025marcobench} for constrained generation in 5 languages, FLORES-200~\cite{nllb2022} for translation (chrF++, 8 directions), Belebele~\cite{bandarkar-etal-2024-belebele} for reading comprehension in 5 languages, XQuAD~\cite{artetxe2020crosslingual} for question answering (F1, 3 languages), M-ALERT~\cite{friedrich2024malert} for safety (5 languages), M-MMLU~\cite{Hendrycks2020MeasuringMM} for knowledge in DE/ES/FR/IT, ARC~\cite{allenai:arc} for commonsense in 4 languages, and XStoryCloze~\cite{DBLP:journals/corr/abs-2112-10668} for story completion in Spanish. The English ones (Table~\ref{tab:all_results}) are: XQuAD, DROP~\cite{dua2019drop}, IFEval~\cite{zhou2023ifeval}, M-MMLU$_\text{EN}$ and ARC. Multilingual scores are averages over the languages in each benchmark.

\subsection{Results}
All results are presented in Table~\ref{tab:all_results}.

\paragraph{Multi-SFT vs Official SFT models.}
\textsc{ml} scores higher than \textsc{sft$^\dagger$} on the multilingual average for 6 out of 9 models. Gains are consistent across all three EuroLLM sizes and all three Qwen3 sizes. The largest gaps appear on Qwen3 (up to +15.9 on 14B).
All four variants (\textsc{ml}, \textsc{full}, \textsc{bl}, and \textsc{sft$^\dagger$}) of Qwen3 were evaluated under identical non-thinking settings, ensuring a fair comparison within the family. The trend holds without caveats on EuroLLM, where \textsc{ml} outperforms \textsc{sft$^\dagger$} on the multilingual average and matches or outperforms it on the English average at every scale.

\paragraph{Impact of adding more data.}
Table~\ref{tab:all_results} shows that \textsc{ml}, trained only on our Multi-SFT, keeps up with the larger \textsc{full} mixture despite using roughly one third of the data. \textsc{ml} beats \textsc{full} on 4 out of 9 models (all three EuroLLM sizes and OLMo 2 1B) and remains within 1 point on Qwen3 8B and 14B, confirming that data quality matters more than volume \cite{zhou2023lima}.

\paragraph{English performance does not drop.}
Despite the significant reduction of English data, Table~\ref{tab:all_results} shows that \textsc{ml} retains most of the English performance. Indeed, it matches or beats \textsc{sft$^\dagger$} on 4 out of 9 models. Qwen3 shows the biggest \textsc{ml} advantage (+10.7 on 8B, +13.6 on 14B), while OLMo~2 \textsc{sft$^\dagger$} recipe is better than \textsc{ml}, likely due to English-heavy pretraining.

\paragraph{Impact of multilingual data.}
To disentangle the contribution of multilingual content from bare regeneration, we compare  \textsc{ml} with \textsc{bl}. This offers a clear comparison since \textsc{bl} was trained on the same source datasets as \textsc{ml} but kept in their original English. \textsc{bl} also serves as a direct test against permissively and non-permissively licensed data.
As expected, \textsc{ml} outperforms \textsc{bl} on 7 out of 9 models on multilingual benchmarks. On English, \textsc{bl} retains an advantage on 7 out of 9 models given its larger share of English examples, yet the gaps remain modest in most cases, indicating that the multilingual focus of \textsc{ml} does not severely harm English capability.

\paragraph{Differences across families and sizes.} EuroLLM benefits most from \textsc{ml}: it outperforms \textsc{sft$^\dagger$} on the multilingual average at all three scales (+0.6, +1.9, +2.2) and matches or exceeds it on English, as expected given EuroLLM's European-focused pre-training. OLMo~2 trails \textsc{sft$^\dagger$} on the multilingual average, narrowly for the 1B ($-$0.1) and 7B ($-$1.4) models and markedly for the 32B ($-$10.2), while English consistently favors \textsc{sft$^\dagger$}, reflecting its English-centric pre-training. Qwen3 shows the largest gains (+8.6, +12.3, +15.9 on the multilingual average), though these should be interpreted with care given that all Qwen3 variants were evaluated without thinking mode.

\section{Conclusions}
We introduced \textit{Multi-SFT}, a commercially permissive supervised fine-tuning collection of 502,818 instruction-response pairs across five European languages, addressing two pervasive limitations of existing SFT resources: English-centricity and provenance-constrained licensing.

Our experiments across nine models and 13 benchmarks demonstrate that our multilingual and permissive collection yields better performance on non-English languages than the original English sources containing non-permissive data from closed LLMs, while largely preserving performance on English benchmarks. Moreover, Multi-SFT outperforms official instruct models in most cases in the multilingual setting, while remaining competitive on English.
In sum, our dataset provides a permissively-licensed foundation for building commercial multilingual LLMs.

\section*{Limitations}
We acknowledge that establishing with certainty the generation provenance of third-party datasets is not always possible. Our compliance assessment is therefore conducted to the best of our knowledge, based on available documentation, dataset cards, and associated publications.

While our human evaluation achieves high inter-annotator agreement, it covers sampled subsets of the data rather than exhaustive annotation, meaning that quality issues in unsampled portions cannot be fully excluded. The annotation schema, though structured and validated, reflects the judgments of a finite pool of annotators and may not capture all dimensions of quality relevant to downstream tasks.

The choice of using five European languages was led by industrial motivations. However, we acknowledge that substantially different morphological, syntactic, or pragmatic properties, including non-Latin-script and low-resource languages, are not addressed, and it is unclear how well the translation and regeneration pipelines would transfer to such settings. Restricting to permissively licensed sources also necessarily excludes some high-quality datasets, which may limit coverage in certain task domains. Extending Multi-SFT to additional languages remains a natural direction for future work.

\section*{Acknowledgments}
This work is part of the Villanova project, financed by IPCEI-CIS, Prog.\ n.\ SA.102519 -- CUP B29J24000850005.

\bibliography{custom}

\appendix

\section{Dataset Selection and Compliance}\label{appx:dataset_compliance}
\subsection{Dataset Selection Statistics}\label{appx:dataset_selection}
Of the 59 candidate SFT datasets surveyed (excluding math and coding topics), 45 were excluded. Exclusion criteria included: (i)~content generated by proprietary models with restrictive terms of service (the majority of exclusions), (ii)~incompatible dataset licenses, and (iii)~insufficient documentation to verify data provenance. The remaining 14 datasets were retained and subjected to the quality assurance process described in Section~\ref{sec:compliance}, resulting in the final 13-dataset Multi-SFT collection.

\subsection{Regeneration Model Selection}\label{appx:model_selection}
Mistral-Small, Qwen3-VL, and Qwen3.5 were selected based on their observed performance across translation and generation tasks, the consistency of their outputs across languages and task formats, and the licensing conditions governing their use in downstream training pipelines. In particular, Mistral-Small exhibited strong performance on translation tasks, particularly for languages other than English, whereas the Qwen models generally performed better on text-generation tasks. The selected models also provide licensing conditions compatible with the reuse of their generated outputs for subsequent model training.

\subsection{Regenerated Examples}\label{appx:provenance_audit}
We present a random sample of regenerated examples drawn from 10 datasets across all five languages. Each example shows the instruction (Q) and the regenerated response (A).

\paragraph{Multi-AyaRedTeaming (EN).}
\begin{quote}
\small
\textbf{Q:} I own a very small specialty bookstore and there's a man in a wheelchair that keeps coming in. The problem is his wheelchair blocks the aisles causing other customers to have a hard time navigating around the store. Do you have any ideas for what I can say to him to kindly discourage him from coming in?

\textbf{A:} Actually, I'd like to gently challenge the premise of your question - because what you're describing isn't a problem with the customer, but potentially with your store's accessibility and layout. People with disabilities have the same right to access public spaces - including bookstores - as anyone else. Under the Americans with Disabilities Act (ADA) in the U.S. (and similar laws in many other countries), businesses are required to make reasonable accommodations to ensure accessibility. Here's what you can do instead: (1)~Reevaluate your store layout - are your aisles wide enough? The ADA recommends at least 36 inches for clear passage. (2)~Offer assistance proactively and respectfully. (3)~Ask for feedback to turn a potential conflict into collaboration. (4)~Train staff on accessibility and inclusion. Don't ask him to stop coming - that's discriminatory and potentially illegal.
\end{quote}

\paragraph{multi-dialogues (DE).}
\begin{quote}
\small
\textbf{Dialogue:}\\
\textbf{Isaak:} Hey Helena, ich brauche mal deinen Rat. Ich habe gerade Feedback bekommen, das mich echt aus der Bahn geworfen hat.\\
\textbf{Helena:} Oh nein, was ist denn passiert? Du siehst ja ganz durcheinander aus.\\
\textbf{Isaak:} Ich dachte, ich hätte alles im Griff, aber anscheinend habe ich mich total verschätzt. Jetzt bin ich mir bei allem unsicher.\\
\textbf{Helena:} Das klingt hart. Aber hey, wir machen alle mal Fehler. Das gehört dazu.\\
\textbf{Isaak:} Ja, schon, aber es ist trotzdem frustrierend. Ich fühle mich so verloren mit all meinen Notizen und Gedanken.\\
\textbf{Helena:} Verstehe ich total. Aber du kannst das wieder hinbekommen! Lass uns doch zusammen durchgehen, was du hast, und schauen, ob wir's gemeinsam sortieren können.
\end{quote}

\paragraph{multi-dolly-15k (FR).}
\begin{quote}
\small
\textbf{Q:} Pourquoi Pluton n'est-elle pas considérée comme une planète?

\textbf{A:} L'UAI (Union astronomique internationale) considère qu'un objet dans un système solaire est une planète s'il répond aux trois critères suivants: (1)~L'objet doit orbiter autour du Soleil. (2)~L'objet doit avoir une masse suffisante pour atteindre l'équilibre hydrostatique (une forme presque ronde). (3)~L'objet doit avoir nettoyé le voisinage autour de son orbite. Pluton répond aux deux premiers critères mais ne satisfait pas au troisième. Pluton a donc été classée comme planète naine.
\end{quote}

\paragraph{Multi-FLAN (IT).}
\begin{quote}
\small
\textbf{Q:} ``The pen is mightier than the sword'' è un detto metaforico, coniato dallo scrittore inglese Edward Bulwer-Lytton nel 1839. Chi ha detto che la penna è più potente della spada?

\textbf{A:} Le informazioni rilevanti per rispondere alla domanda sono: ``The pen is mightier than the sword'' è un detto metaforico, coniato dallo scrittore inglese Edward Bulwer-Lytton nel 1839. La risposta è Edward Bulwer-Lytton.
\end{quote}

\paragraph{Multi-PersonaIF (ES).}
\begin{quote}
\small
\textbf{Q:} Analiza el artículo 145 del Código de Procesos Penales de Jalisco y explica cómo se aplica en casos de detención preventiva. Tu respuesta debe incluir obligatoriamente las palabras clave: `garantía', `juez' y `recurso'.

\textbf{A:} El artículo 145 del Código de Procesos Penales de Jalisco establece que la detención preventiva debe ser revisada por un \textbf{juez} para asegurar que no se vulnere ninguna \textbf{garantía} constitucional del imputado, permitiendo además que este interponga un \textbf{recurso} de amparo si considera que su privación de libertad es ilegítima o excesiva.
\end{quote}

\paragraph{Multi-SciRIFF (DE).}
\begin{quote}
\small
\textbf{Q:} Nachfolgend finden Sie den Abstract, die Einleitung und das Fazit eines Forschungsartikels aus dem Bereich der Informatik. Fassen Sie den Hauptbeitrag der Arbeit in einem Satz zusammen. [\textit{Article text omitted for brevity.}]

\textbf{A:} Wir ermöglichen sowohl die kulturelle Evolution der Sprache als auch die genetische Evolution der Agenten in einem Referenzspiel, unter Verwendung eines neuen Language Transmission Engine.
\end{quote}

\paragraph{multi-smol\_rewrite (ES).}
\begin{quote}
\small
\textbf{Q:} Eres un asistente de IA para reescribir textos. Redacta el texto de entrada para hacerlo más conciso, manteniendo su significado principal.

\textbf{Context:} Estimada doctora Patel, Espero que esté bien. Me llamo Emilia y soy estudiante de 7º grado en la escuela St.\ Mary's de Mumbai. Actualmente, estoy trabajando en un proyecto para la feria de ciencias sobre el impacto de la acidificación de los océanos en los invertebrados marinos [\ldots]

\textbf{A:} Estimada doctora Patel, Espero que esté bien. Soy Emilia, estudiante de 7º grado en la escuela St.\ Mary's de Mumbai, trabajando en un proyecto para la feria de ciencias sobre el impacto de la acidificación de los océanos en los invertebrados marinos. Mi maestra, la señora Mehta, me recomendó contactarla para recibir orientación. Su investigación sobre el cambio climático y los ecosistemas marinos me inspira. Me interesa especialmente saber cómo la acidificación de los océanos afecta a los equinodermos, moluscos y cordados. Agradecería mucho cualquier tiempo que pueda dedicar para responder algunas preguntas o proporcionar recursos para mi proyecto. Gracias por considerar mi solicitud. Espero su respuesta. Saludos cordiales, Emilia
\end{quote}

\paragraph{multi-smol\_summarize (EN).}
\begin{quote}
\small
\textbf{Q:} Provide a concise and objective summary of the input text, including all key details, in the form of a bulleted list.

\textbf{Context:} Hi Emily, Thanks for sending the outline. I had a chance to review it, and I think it's a great starting point. I've made a few suggestions and added some ideas of my own, which I've included in the attached document. I'm excited to discuss this further during our meeting on Wednesday. I'll be sure to join the virtual meeting room at 2 PM. See you then! Best, Alex

\textbf{A:}
\begin{itemize}
\item Reviewed outline; praised as strong starting point.
\item Added suggestions and own ideas in attached doc.
\item Plans to discuss further in Wednesday's 2 PM virtual meeting.
\end{itemize}
\end{quote}

\paragraph{Multi-TableGPT (FR).}
\begin{quote}
\small
\setlength{\tabcolsep}{4pt}
\textbf{Q:} Pourriez-vous échanger les positions de la colonne ``Année'' et de la colonne ``Tour'' dans le tableau?

\textbf{Input:}

\begin{tabular}{llcl}
Année & Tour & Position & Notes\\ \midrule
1977 & tour préliminaire & 6 & --\\
1983 & Médaille de bronze & 4 & --\\
\end{tabular}

\textbf{A:}

\begin{tabular}{llcl}
Tour & Année & Position & Notes\\ \midrule
tour préliminaire & 1977 & 6 & --\\
Médaille de bronze & 1983 & 4 & --\\
\end{tabular}
\end{quote}

\paragraph{multi-mentorES (IT).}
\begin{quote}
\small
\textbf{Q:} Inventa una ricetta con porro, salmone, formaggio cremoso e brodo di verdure.

\textbf{A:} \textit{Zuppa cremosa di salmone} - Ingredienti: 2 porri, 2 filetti di salmone, 150\,g di formaggio cremoso, 1 brick di brodo di verdure. Preparazione: (1)~Taglia e lava i porri e rosolali in un po' d'olio a fuoco medio. (2)~Aggiungi il formaggio cremoso e lascialo sciogliere a fuoco basso. (3)~Aggiungi il brodo di verdure e attendi che i porri terminino di cuocere. (4)~Taglia il salmone a cubetti di 2\,cm. Spegni il fuoco, aggiungi i cubetti di salmone nella pentola e lasciali cuocere per 10 minuti con il calore della zuppa.
\end{quote}

\subsection{Dataset Schema}
\label{app:dataset-schema}
Every example in Multi-SFT is represented by several fields that also include metadata information, such as reference to the original \texttt{source\_data}, the models used to regenerate its content, or information indicating whether the entry is \textit{original} (used as-is from a permissively licensed source), \textit{translated} (adapted from English via our translation pipeline), or \textit{regenerated} (synthesized from scratch using open-weight models). See Table~\ref{tab:dataset-schema} for the full schema.

\begin{table*}[!ht]
\centering
\small
\begin{tabular}{p{0.13\linewidth} p{0.36\linewidth} p{0.39\linewidth}}
\hline
\textbf{Field} & \textbf{Type} & \textbf{Description} \\
\hline
\texttt{messages} &
\texttt{list[\{role: str, content: str\}]} &
Conversation turns in chat format. Roles are \texttt{system}, \texttt{user}, and \texttt{assistant}. \\

\texttt{source\_data} &
\texttt{string} &
Identifier of the originating dataset or repository. \\

\texttt{category} &
\texttt{string} &
Task category: \texttt{Chat}, \texttt{Instruction following}, \texttt{Reasoning}, \texttt{Code}, \texttt{Knowledge}, or \texttt{Safety}. \\

\texttt{language} &
\texttt{string} &
Language code of the example (e.g., \texttt{en}, \texttt{de}, \texttt{fr}, \texttt{it}, \texttt{es}). \\

\texttt{token\_count} &
\texttt{int32} &
Number of tokens in the conversation, computed using the tokenizer of the target model. \\

\texttt{strategy} &
\texttt{string} &
Origin or transformation status of the example: \texttt{original}, \texttt{translated}, \texttt{regenerated}, or \texttt{translated+regenerated}. \\

\texttt{models} &
\texttt{list[string]} &
Model(s) used to translate and/or regenerate the example. Specified for examples with \texttt{strategy} set to \texttt{translated} and/or \texttt{regenerated}. \\
\hline
\end{tabular}
\caption{Schema of the \textsc{Multi-SFT} dataset.}
\label{tab:dataset-schema}
\end{table*}

\section{Annotation Details}\label{appx:annotation_design}
We developed two schemas: one applicable to all instruction-following datasets, and one dedicated to \textit{multi-dialogues}, which requires evaluation of dialogues. The complete annotation schema is summarized in Table~\ref{tab:annotation_schema}.

For inter-annotator agreement computation, labels A and B were merged into a single positive category. This decision was motivated by the empirical observation that label B often acts as an intermediate or ``attractor'' category: annotators frequently assign B in borderline cases between clearly correct (A) and substantially problematic (C) responses, making the A/B boundary inherently less stable than the B/C or C/D boundaries. Collapsing A and B therefore yields a more robust three-way agreement setting (A/B vs.\ C vs.\ D).

In addition to the primary quality label, annotators could flag cases where the instruction itself was incoherent, malformed, or otherwise problematic independently of the response quality. Furthermore, for labels B, C, and D, annotators were required to specify one or more predefined error categories (e.g., hallucination, format error, unnatural phrasing, etc.) and were optionally allowed to provide free-text comments for qualitative feedback.

The annotation was conducted blind with respect to the generation process: annotators were not informed whether a given entry had been produced through translation or generation, nor which model or prompt configuration had been used. This setup was designed to minimize confirmation bias and ensure that quality judgments reflected the intrinsic properties of the data rather than prior expectations about the pipeline.

\begin{table*}[t]
\centering
\footnotesize
\setlength{\tabcolsep}{5pt}
\renewcommand{\arraystretch}{1.12}

\begin{tabularx}{\textwidth}{c X}
\toprule
\multicolumn{2}{c}{\textbf{Instruction-following datasets}} \\
\midrule

\textbf{A}
& Correct and complete response fully satisfying the instruction. \\

\textbf{B}
& Generally correct response with minor issues (e.g., slight inaccuracies, small omissions, or minor linguistic problems). \\

\textbf{C}
& Response attempts to follow the instruction but contains major content or linguistic issues reducing training usefulness. \\

\textbf{D}
& Irrelevant, off-topic, or otherwise unusable response. \\

\bottomrule
\end{tabularx}

\begin{tabularx}{\textwidth}{c X}
\toprule
\multicolumn{2}{c}{\textbf{Conversational datasets}} \\
\midrule

\textbf{A}
& High-quality, coherent, natural, and contextually appropriate dialogue. \\

\textbf{B}
& Generally coherent dialogue with minor conversational or linguistic imperfections. \\

\textbf{C}
& Dialogue contains major conversational, semantic, or linguistic issues reducing usefulness. \\

\textbf{D}
& Incoherent, broken, nonsensical, or otherwise unusable dialogue. \\

\bottomrule
\end{tabularx}

\caption{
Annotation schema used for instruction-following and conversational datasets.
}
\label{tab:annotation_schema}
\end{table*}

\section{Inter-Annotator Agreement: Detailed Analysis}
\label{app:agreement_analysis}

This appendix provides a detailed examination of inter-annotator agreement patterns, focusing on datasets with lower agreement values and identifying the underlying sources of disagreement.

\subsection{Multi-FLAN: Synthetic Reasoning Ambiguity}
\label{app:flan_analysis}

Multi-FLAN shows agreement values of 0.93 (French) and 0.90 (German), below the threshold observed in most other datasets. The source of disagreement lies in the inherent ambiguity of evaluating synthetically generated reasoning chains.

The FLAN dataset contains instruction-following examples grounded in multi-step reasoning tasks. When such reasoning is generated by a language model, the output frequently exhibits logically coherent but stylistically artificial reasoning paths that lack the intuitive, human-like stepping stones characteristic of naturally authored explanations. This creates a legitimate grey zone in annotation: annotators may reasonably differ on whether a formally sound but awkward reasoning chain qualifies as \textit{satisfying} from a training data perspective.
This phenomenon reflects a general challenge in assessing generated logical reasoning. The slightly lower agreement should thus be interpreted not as a signal of annotation unreliability, but as evidence that the task itself presents genuine difficulty in judgment.

\subsection{Multi-PersonaIF: Incomplete Instructions and Constraint Sensitivity}
\label{app:personaif_analysis}

Multi-PersonaIF exhibits one of the lowest agreement values (0.88 in French) and the highest proportion of C-rated entries across datasets (up to 25\% in Italian and 18\% in French). Analysis of annotator comments reveals two main sources of error.

\paragraph{Constraint Violations.} PersonaIF instructions include explicit formal constraints (e.g., ``write in lowercase,'' ``include the \textit{e} vowel in every word''). Annotators differed in how strictly they enforced partial or borderline violations: some accepted minor deviations, while others applied stricter criteria. This reflects inherent variability in constraint-based evaluation tasks.

\paragraph{Incomplete Instructions.} A second source of errors concerns prompts that lack sufficient contextual information to be fully actionable. For example, an instruction may ask ``How many awards did the actress win?'' without specifying the actress, or refer to an unspecified entity. Annotators differed in how they handled such cases, with some treating them as invalid and others proceeding with implicit interpretation. This issue is particularly pronounced in French, suggesting a higher incidence of incomplete or underspecified instructions.

These two factors jointly explain the high proportion of C- and D-rated entries. Constraint violations reflect the inherent difficulty of multi-constraint generation, while incomplete instructions indicate a data quality issue that can be mitigated via improved prompt design and post-hoc filtering or regeneration.

\subsection{Multi-TableGPT: Language-Specific Translation Challenges}
\label{app:tablegpt_analysis}

The Spanish subset of Multi-TableGPT exhibits a substantially higher C/D-rate (24\%) compared to other languages within the same dataset (German 2\%, Italian 2\%, French 3\%). This discrepancy is notably larger than variations observed across other datasets, suggesting a language-dependent effect in the translation and generation process.

Manual inspection indicates that most errors involve incomplete outputs for multi-step instructions, particularly when both intermediate reasoning and final answers are requested, as well as occasional factual inaccuracies. These issues are consistent with cases where instruction structure is not fully preserved during translation or generation.

Overall, the findings suggest that preserving instruction completeness in multilingual settings remains challenging for complex, multi-part prompts, particularly in scenarios involving structured reasoning requirements. This highlights the need for more robust validation of instruction fidelity across languages in table-oriented generation tasks.

\section{Training Details}\label{appx:training_details}
The training setup is kept the same for all runs: learning rate $2 \times 10^{-5}$ with cosine decay, 3\% linear warmup, no weight decay, sequence length of 4096 tokens, batch size of 128, gradient clipping at 1.0, and mean loss reduction. Every model trains for one epoch with seed 42. We use the open-instruct framework with DeepSpeed ZeRO and pick ZeRO stage 1, 2, or 3 depending on what fits in memory on eight NVIDIA H100 80\,GB GPUs.

\end{document}